\documentclass[journal]{IEEEtran}
\IEEEoverridecommandlockouts

\usepackage{cite}
\usepackage{amsmath,amssymb,amsfonts}
\usepackage{algorithm}      
\usepackage{algpseudocode} 
\usepackage{graphicx}
\usepackage{textcomp}
\usepackage{xcolor}
\usepackage{url}
\usepackage{array}
\usepackage{booktabs}
\usepackage{multirow}
\usepackage{subcaption}
\usepackage{enumitem}
\usepackage{fontawesome5}
\usepackage{xspace}
\usepackage{hyperref}
\usepackage{soul}
\usepackage{hyperref}
\usepackage{xr-hyper}

\sethlcolor{yellow}

\def\BibTeX{{\rm B\kern-.05em{\sc i\kern-.025em b}\kern-.08em
    T\kern-.1667em\lower.7ex\hbox{E}\kern-.125emX}}

\newcommand{\sys}{Neurosymbolic Deep Analyst\xspace}       
\newcommand{\deepanalyst}{Deep Analyst\xspace}             
\newcommand{\querier}{EXASAGE Reflect Agent\xspace}        
\newcommand{\exao}{EXASAGE Query Tool\xspace}              
\newcommand{\datalaketool}{Datalake Query Tool\xspace}     
\newcommand{\dr}{Deep Neuro-Symbolic Data Retriever\xspace}
\newcommand{\codeagent}{Deep Code Agent\xspace}            

\begin{document}

\title{Symbolic Separation: Grounding Deep Agents in\\
Knowledge Graphs for Trustworthy Operational Data Analytics}

\author{Baibek Davletiyarov, Junaid Ahmed Khan, and~Andrea Bartolini%
\thanks{The authors are affiliated with the DEI Department, University of Bologna, Italy.
E-mails: \texttt{\{baibek.davletiyarov2, junaidahmed.khan, a.bartolini\}@unibo.it}.}%

}

\markboth{IEEE Transactions on Knowledge and Data Engineering -- Special Issue on Data and Knowledge Empowered Generative AI}%
{Khan \MakeLowercase{\textit{et al.}}: Grounding Deep Agents in Knowledge Graphs for Trustworthy Operational Data Analytics}

\maketitle

\begin{abstract}
Generative AI promises natural-language access to the massive \emph{numerical}
telemetry of data centers and Industry~4.0 installations, yet text-to-query and
tool-using agents stay unreliable: even frontier models answer little more than
half of real-world database questions, and far fewer of the multi-step,
operational ones, because the LLM must \emph{compose} how heterogeneous sources relate and
hallucinates the relations, not just the fields. We propose \emph{symbolic separation}: a deep agent reasons freely
but may \emph{act} on data only through an ontology-constrained Virtual Knowledge
Graph with deterministic pre-execution validation. Unlike a tool API's
\emph{interface contract}, this \emph{domain-semantic contract} turns a complex
question into one validated graph traversal instead of LLM-inferred joins.
Instantiated as the \sys{} and evaluated on 49.9\,TB of supercomputer telemetry
against a rigid-workflow and a non-symbolic ablation, it raises end-to-end task
success from 43\% to 86\%, prevents silent data-integrity errors that no syntactic
check catches, and cuts token cost by $2.4\times$, letting a smaller on-premise model
outperform a larger one.
\end{abstract}

\begin{IEEEkeywords}
Neurosymbolic AI, Deep Agents, Knowledge Graphs, Virtual Knowledge Graph,
Ontology-Based Data Access, Ontological Constraints, LLM Agents, Model Context
Protocol, Operational Data Analytics, Trustworthy AI, Hallucination Mitigation.
\end{IEEEkeywords}

\section{Introduction}\label{sec:intro}
Over the last two decades, digital transformation has increasingly focused on instrumenting, sensing, and connecting the physical world. The Internet of Things (IoT) paradigm envisions a worldwide network of interconnected smart physical entities that continuously generate large volumes of operational data \cite{siow2018analytics}. In industrial contexts, this trajectory was consolidated under the Industry 4.0 paradigm, where cyber-physical systems, industrial IoT, and data-driven analytics became central to monitoring, understanding, and optimizing production processes \cite{Duan2024}. As a result, the last decade has seen a shift from merely collecting industrial data to extracting actionable knowledge from large-scale, heterogeneous, and streaming data sources \cite{mohammadi2018deep}. However, the discovery, modeling, and optimization of complex processes have remained constrained by the complexity of data-science workflows and the availability of expert data scientists and domain specialists \cite{debie2022automating}. Extracting value from data-lakes and IoT installations traditionally imposes a three-fold barrier on the user, who must simultaneously be a domain expert, an expert in the specific monitoring deployment, and an expert in the storage backend's query language and API.

LLMs appear to be the natural interface to lower this barrier, but purely
probabilistic models are unreliable on exactly the dimensions that matter for
mission-critical operations: they hallucinate non-existent entities and
relations~\cite{haluc_survey} and cannot track private, installation-specific
schemas. The canonical attempt to bridge language and data -- text-to-databases
-- remains an open problem. On BIRD, a benchmark of $12{,}751$
questions over $95$ real databases, even GPT-4 reaches only $40.08\%$ execution
accuracy against $92.96\%$ for human experts~\cite{LLM-SQL-real-world}; 
On dynamic, multi-turn workloads, GPT-4o achieves ($58.34\%$) overall turn-level accuracy, but only ($23.81\%$) under the benchmark’s stricter task-level pass@5 criterion~\cite{sun2026dysqlbench}. This gap suggests that partial success on individual turns does not reliably translate into robust completion of an evolving interaction. Moreover, residual errors are predominantly semantic rather than syntactic: NL2SQL-BUGs catalogues (9) major categories and (31) subcategories of semantic errors in text-to-SQL generation~\cite{liu2025nl2sqlbugs}. LLMs achieve only 75.16\% accuracy when auditing these logical flaws. Yet, when analyzing human-verified annotations, this auditing approach revealed that 6.91\% of BIRD’s ground-truth queries contain undetected semantic errors. Therefore, overcoming the text-to-database bottleneck requires moving beyond raw execution accuracy toward frameworks capable of deep semantic verification.

The difficulty is sharpest on \emph{numerical
operational telemetry}: state-of-the-art timeseries data agents answer $\sim$73\%
of stateless queries but only $\sim$34\% of stateful and $\sim$10\% of incident
(anomaly) queries, with failures dominated by schema confusion and wrong
table/column selection~\cite{maddi2026agentfuel}. The same pattern appears when an
LLM writes analysis \emph{code} directly over IoT sensor data: in a published
breakdown of its failures, failed data imports, mishandled datetime formats, and
wrong column names together account for over $80\%$ of the
errors~\cite{zong2024integratinglargelanguagemodels}. Three difficulties compound:
(i)~connecting to the data lake and writing valid queries, (ii)~knowing which
sensor or metric captures a given physical effect on which component, and
(iii)~combining multiple sources -- that is, knowing and traversing the relations
between the data and the observed concepts.

A promising remedy is to ground
generative models in \emph{structured knowledge}~\cite{pan2024unifying}. Knowledge Graphs (KGs) expressed in Resource Description Framework (RDF) unify heterogeneous sources under a shared, ontology-defined schema and expose them through expressive query languages such as SPARQL.  Our own prior work demonstrated this for data center telemetry: ExaQuery~\cite{khanExaQueryProvingData2024} introduced a
domain ontology for operational data, and on top of it the EXASAGE framework~\cite{khan_fgcs_exasage} -- which we hereafter call \exao -- translated
natural language queries to SPARQL resolved against a \emph{Virtual} Knowledge Graph (VKG)~\cite{vkg_overview}---a dynamic graph built on demand, specific to the user request---evaluated on 1K complex query randomly generated from ten archetypes. On those EXASAGE 
reached $93.6\%$ answer accuracy versus $25\%$ for direct LLM-to-NoSQL
generation. Furthermore, by avoiding complete graph materialisation, this virtual approach entirely eliminates storage explosions while maintaining low latency. Crucially, the VKG already moved the
construction of cross-source relationships upstream into a connected graph, so that a question is answered by traversing pre-existing edges rather than by the LLM inferring joins -- a property we make central in this paper.

\smallskip
Despite this accuracy, EXASAGE~\cite{khan_fgcs_exasage} is a \emph{rigid, single-shot workflow}. Entity
extraction relies on hand-written regular expressions over a fixed category set
(\texttt{node}, \texttt{rack}, \texttt{job}, \texttt{metric}, \texttt{plugin},
plus temporal markers), built around the ten query archetypes pattern only. This limits the scope of the answerable queries and prevents generalisation: a conceptual term such as ``power consumption'' fails to map to the right metric if it is not literally present in the plugin--metric table, and temporal phrasings such as ``last 10 hours'' are dropped because they do not match the expected
\texttt{[YYYY-MM-DD HH:MM:SS]} format. Furthermore, the workflow has no recovery mechanism. Although a final validation step monitors the graph database endpoint for runtime execution errors, this pass only catches syntax failures that cause query execution to fail. It cannot detect semantic errors or verify ontology conformance; thus, logically flawed but syntactically correct queries execute silently and return incorrect data. Crucially, there is no loop in which the system can reflect, re-plan, or ask for clarification. Even though its accuracy surpassed direct LLM-to-NoSQL, and it is to the best of the author's knowledge the only work in SoA targeting ontology and VKG grounding of timeseries data NL2SQL, EXASAGE~\cite{khan_fgcs_exasage} needs to be validated out-of-design-samples to validate its design efficacy.


At the same time, we are facing a broader shift in GenAI from externally orchestrated LLM
workflows to long-horizon \emph{agentic} systems -- and, increasingly, \emph{deep
agents} that plan, call tools, delegate to sub-agents, and revise their behaviour over multi-step tasks~\cite{yao2023react, shinn2023reflexion, langchain_deepagents}.

For operational analytics, however, this added autonomy helps only if the agent's actions are grounded: without a symbolic contract over entities, metrics, relations, and admissible queries, a more capable agent merely gains more ways to hallucinate over the infrastructure it controls. This leads directly to the two research questions addressed in this manuscript:
\begin{description}
    \item[RQ1:] Is encapsulating EXASAGE~\cite{khan_fgcs_exasage} as a tool within a modern deep agent design sufficient to absorb the rigidity of its single-shot workflow, or must its internal architecture be entirely rethought to answer generalized queries?
    \item[RQ2:] Are deep agents powered by modern LLMs capable of navigating real-world time-series databases on their own, or do they fundamentally require a symbolic representation of the queried data?
\end{description}

To answer these research questions, we introduce the \emph{symbolic separation} design concept for data deep agent:  
the agent reasons freely in the neural layer, but
it can only act on data through a symbolic layer -- an
ontology-constrained knowledge graph -- that validates each access against the schema before it reaches the data.

Mediating data access through an ontology is not itself new, and existing LLM$+$KG
integrations span a spectrum~\cite{pan2024unifying, ma2025llmkgqa}: from
\emph{verbalising} retrieved triples into the prompt (KG-RAG), through giving the
agent a \emph{tool or semantic-layer API} over the data, to having the LLM
\emph{author formal queries} (text-to-SQL/SPARQL). In every case, however, the LLM still composes the \emph{relational structure} of a complex answer -- it issues
multiple calls or joins and infers how their results relate -- and that cross-call composition is exactly where schema hallucination concentrates and worsens with
complexity~\cite{LLM-SQL-real-world}. A tool/semantic-layer agent enforces only an \emph{interface contract}: each individual call is vocabulary- and type-valid, but the relations between calls are not. Ontology-Based
Data Access and Virtual Knowledge Graphs~\cite{poggi2008linking, xiao2018obda} are the exception -- access \emph{is} the ontology -- yet no prior system connects them with deterministic, pre-execution validation inside a
self-correcting \emph{deep agent}, and least of all for timeseries data and operational
telemetry. In contrast, the proposed symbolic separation enforces a
\emph{domain-semantic contract}: because relationships are materialised and
ontology-validated in the VKG before any query is issued, a complex question
becomes a single traversal over pre-existing, validated edges, and the agent is structurally prevented not just from naming a non-existent field but from inventing how data relates, 
leading to a trustworthy data analyst agent.

To validate it, in this manuscript we propose a \emph{deep neuro-symbolic analyst} framework and evaluate it on two sets of queries in a large corpus of real data center telemetry data. The two set of queries stress one complex multi-turn data analysis query involving data retrieval, statistical post-processing and data visualization tasks, and one data retrieval-only queries. It is a multi-agent system: a deep-agent coordinator and two deep sub-agents -- a \dr{} and a \codeagent. The \dr{} integrates a newly proposed \querier{} with a deep sub-agent via MCP tool calling for long-horizon planning
: the LLM performs only entity extraction and
SPARQL generation, using the domain ontology as context, while input validation,
data retrieval, virtual knowledge-graph creation, and query resolution are handled
by trustworthy, auditable symbolic components. The \codeagent{} runs statistical
analysis and visualisation in a sandbox. This realises symbolic separation: the
coordinator reasons freely on text, while data are queried only through the
neuro-symbolic \querier. Our result shows that:

\begin{itemize}[leftmargin=1.4em]
  \item On end-to-end analytics queries (data retrieval + code execution), the proposed \sys{} with Qwen3.6-35B-A3B achieves 86\% success, compared to 7\% for the EXASAGE\cite{khan_fgcs_exasage} SoA baseline and 43\% for a non-symbolic deep analyst.
  \item On retrieval-only queries only from HPC operational data analytics studies, the proposed approach outperforms SoA baselines and achieves the 88\% of accuracy.
  \item The proposed \querier{} consumes 2.4$\times$ lower median tokens over the best alternative configuration, produces zero hallucinated outputs, and exhibits a fail-fast retry behaviour that avoids wasteful computation.
  \item These results confirm that symbolic separation -- decomposing symbolic pipelines into focused, specialised sub-agents -- not adding an agentic loop to a monolithic prompt, is the key to accurate and trustworthy analytics agents.
\end{itemize}

The remainder is organised as follows. Section~\ref{sec:soa} reviews the state
of the art. Section~\ref{sec:method} details the approach -- the reference
architecture, the symbolic core, and the \sys use-case implementation -- and the
comparison architectures. Section~\ref{sec:results} presents the evaluation, and
Section~\ref{sec:concl} concludes.

\section{State of the Art}\label{sec:soa}
\subsection{Operational Data Analytics for Data Centers}
Modern data centers and HPC systems are complex industrial plants instrumented with millions of sensors~\cite{netti2021conceptual, ODA}. State-of-the-art
telemetry frameworks, often called Operational Data Analytics
(ODA)~\cite{netti2022operational}, rely on NoSQL stores to absorb heterogeneous, schema-less streams at scale~\cite{CACM-NoSQL}, shifting the burden of establishing relationships between sources onto the user~\cite{querying_heterogeneous_nosql, scherzinger2013managingschemaevolutionnosql}.
The M100~ExaData campaign~\cite{m100nature} made the largest such corpus public: including management, workload, facility, and infrastructure data from all $980{+}$ compute nodes over two and a half years --
$49.9$\,TB uncompressed, the largest public supercomputer dataset to
date~\cite{m100nature} -- later partitioned as Parquet across nine plugins (IPMI,
Ganglia, Vertiv, Schneider, Logics, Weather, Nagios, SLURM, and the Job table),
each with a distinct, sometimes inconsistent, schema. The dataset has supported
thermal-hazard prediction, anomaly detection, and predictive
maintenance~\cite{AksarTPDS_anomalies_at_runtime, borghesiADAP}, but each
analysis still demands deep query expertise.

\subsection{Knowledge Graphs for Heterogeneous Telemetry}
KGs address heterogeneity by integrating diverse sources into a unified, ontology-defined semantic framework expressed as RDF subject--predicate--object triples~\cite{ji2021survey, rdf_soa1}, and have been applied across IoT domains
for recommendation, security, middleware, and fault
diagnosis~\cite{Iot_KG_useCase1, iotKGMiddleware, knowledge_based_diagnosis_iiot}.
For data-center telemetry, ExaQuery~\cite{khanExaQueryProvingData2024} proposed a
domain ontology in which vertices are measurements and edges are topological and compositional relationships. Materialising a full KG for telemetry is
prohibitively expensive in storage: one month of data for just one data collector moves from $4.00$
GiB in Parquet-based NoSQL format to ${\sim}3$ TiB when materialized in a KG, approximately $745\times$ of storage increase; the \emph{Virtual} Knowledge Graph paradigm~\cite{vkg_overview} avoids this by constructing, per query, only the
sub-graph required. In EXASAGE~\cite{khan_fgcs_exasage} the VKG approach led to a maximum of $0.17$ GiB storage size across all the evaluated queries.

\subsection{LLMs $+$ Knowledge Graphs Taxonomy}
\label{sec:soa-tax}
A large body of literature now combines LLMs with structured knowledge. The TKDE roadmap of Pan et al.~\cite{pan2024unifying} organises it by integration direction (KG-enhanced LLMs, LLM-augmented KGs, synergised). Under this taxonomy EXASAGE would fall into the LLM-augmented KGs, but applied to timeseries data. In the following, we will analyse the surveyed approaches based on \emph{whether data access is mediated and validated by the symbolic layer}. In particular, from \emph{where the relational structure of a complex answer came from} -- materialised in the representation, or inferred by the LLM at query time. 
\\ 
\paragraph{KG-augmented generation (KG-RAG) and KG-guided reasoning:}
KAPING and related methods retrieve triples and verbalise them into the prompt; the LLM then free-generates the answer
text~\cite{baek-etal-2023-knowledge, agrawal2024kghalluc}. The KG informs but
does not gate the output, so hallucination persists at generation time.
Similarly, Think-on-Graph~\cite{sun2024thinkongraph} and Reasoning-on-Graphs~\cite{luo2024rog} let the LLM traverse the
KG during the thinking process, thereby improving faithfulness, but the final answer is still LLM-generated free text grounded by retrieved paths -- the
relational composition of the answer is still the model's. Hence,
\emph{no symbolic separation.}
\paragraph{Semantic parsing: text-to-SQL}
Here the LLM's act is to emit a formal query~\cite{hong2025text2sqlsurvey}. 
The LLM must \emph{author the joins}, and there is usually no deterministic ontology validator rejecting hallucinated schema elements before execution; schema/join hallucination is a documented, persistent failure that grows with query complexity -- on the BIRD real-database benchmark even GPT-4 attains only $40.08\%$ execution accuracy against $92.96\%$ for human experts~\cite{LLM-SQL-real-world}. To counter surface-level formatting failures, \emph{grammar-constrained decoding (GCD)} techniques \cite{geng2023grammar} have been used that force token generation to strictly adhere to formal SQL context-free grammars or regular expressions at inference time. However, GCD operates purely at the lexical and syntactic level; it guarantees valid keywords and balanced parentheses, but remains blind to the underlying schema or domain invariants. Consequently, it cannot prevent the generation of syntactically flawless queries that nonetheless reference non-existent tables or violate relational constraints. This remains an unsettled problem: a recent state-of-the-art survey lists trustworthy/interpretable, interactive, and multi-database NL2SQL among the field's central \emph{open} problems~\cite{luo2025nl2sqlsota, floratou2024nl2sqlnot}, a dedicated benchmark classifies NL2SQL semantic errors into $9$ categories and $31$ subcategories~\cite{liu2025nl2sqlbugs}, and on realistic multi-turn, state-altering workloads even strong models degrade sharply -- GPT-4o reaches only $58.34\%$ overall and $23.81\%$ under a strict pass@5, with the dominant bottleneck being \emph{intent understanding and state planning rather than surface syntax}~\cite{sun2026dysqlbench}. Hence, \emph{no symbolic separation.}
\paragraph{Semantic layers / tool-use agents}
A widespread pattern gives an LLM agent a semantic layer or ontology-mapped tool API over the data~\cite{agenticrag2025sok, yao2023react, schick2023toolformer}. This enforces an \emph{interface contract} -- each call is vocabulary-and type-valid -- but a complex question requires multiple calls whose results
the LLM must relate itself, and that cross-call relational logic is where hallucination concentrates; empirically, such agents over raw timeseries tables fail most on the queries that demand multi-source composition and state~\cite{maddi2026agentfuel}. Hence, \emph{medium symbolic separation.}
\paragraph{Ontology-Based Data Access and Virtual Knowledge Graphs}
OBDA/VKG~\cite{poggi2008linking, xiao2018obda, calvanese2017ontop, vkg_overview}\ is, by construction, ``all data access goes through the ontology'': the ontology is the sole query vocabulary and relationships are declared once, not inferred per query. Recent works in the domain of Knowledge Graph Question Answering (KGQA) implement LLM-Augmented KG Question Answering~\cite{pan2024unifying}. The authors of SPINACH~\cite{liu2024spinach} present an AI agent that generates SPARQL queries over a KG, improving accuracy from $3.9\%$ to $21.4\%$ on a newly proposed dataset. Its error analysis attributes $40\%$ of failures to fetching or misusing the wrong \emph{property/relation} and $30\%$ to an inability to compose sufficiently complex SPARQL ($\sim$$70\%$ together), against only $5\%$ from surface formatting -- that is, the dominant failures are \emph{relational and compositional}, not syntactic. When it comes to
\emph{numerical} timeseries and sensor telemetry KGQA, VKG become necessary and only EXASAGE~\cite{khan_fgcs_exasage} and its VKG-chatbot
extension~\cite{khan2025datacenteriottelemetry} is present as an approach in the literature. The reported accuracy is $93.6\%$ for correctly generated and executed graph queries, compared to only $25\%$ accuracy for standard NoSQL query generation. This evaluation is conducted on 1K queries randomly generated from ten archetypes ones. This limits the generalization of the approach in end-to-end data analyst tasks. Hence, \emph{high symbolic separation.} 
 
\smallskip
Furthermore, recent evaluation on conversational timeseries analytics over IoT, observability, and telecom telemetry, state-of-the-art data
agents answer $\sim$73\% of simple (stateless) queries but only $\sim$34\% of
stateful and $\sim$10\% of incident (anomaly-detection) queries, and their
failures are dominated by schema confusion, wrong time-window selection, and
data-unaware baselines rather than by malformed
syntax~\cite{maddi2026agentfuel}. 

These are diagnostic \emph{benchmarks and evaluation frameworks}, not grounded systems; 
the present paper targets the same gap with a symbolic-separation \emph{architecture} that removes the schema- and join-level failure modes they expose. 

Agentic patterns underlie our realisation: ReAct interleaves reasoning with tool
calls~\cite{yao2023react}, Reflexion adds verbal self-reflection~\cite{shinn2023reflexion},
and \emph{deep agents} add a planner/orchestrator that delegates to focused
subagents and skills~\cite{langchain_deepagents}, with capabilities exposed
through MCP~\cite{mcp_spec, fastmcp}. At the model level, tool use is increasingly
first-class -- Qwen3.6-35B-A3B and GPT-OSS-120B expose native function calling and agentic
execution~\cite{qwen2026functioncalling, openai2025gptoss} and coding agents such
as SWE-agent operate through an agent--computer
interface~\cite{yang2024sweagent} -- which is what makes a self-hostable deep agent
over telemetry practical. However, none of the work in the literature applies the symbolic separation approach to timeseries, IoT data. 

In contrast, in this paper we propose: 
(i)~naming symbolic separation and drawing the
\emph{interface-vs domain-semantic-contract} distinction that explains why
pre-connected, ontology-validated relations beat query-time join inference on
complex questions; (ii)~a \emph{deterministic ontology-conformance validator}
that rejects classes/properties/domain-range violations before execution --
strictly stronger than \exao's syntactic regex refinement and than grammar-only
constrained decoding; (iii)~realising all this inside a \emph{deep agent}
(multi-step plan--act--observe, reflection, clarification, sandboxed
computation) rather than a single-shot workflow; and (iv)~doing so for
\emph{numerical operational telemetry}, the domain where prior work is sparsest. We therefore make no claim of being ``first to mediate access
through an ontology''; we claim the first \emph{trustworthiness-framed,
deep-agent, deterministically-validated} instantiation of symbolic separation
for operational data analytics.

\section{Methodology}\label{sec:method}
We present the approach top-down. Section~\ref{sec:arch} gives the overall
architecture -- the flow of information, the role of each component, and the
symbolic-separation boundary that makes the design trustworthy -- instantiated as
\sys. Section~\ref{sec:subagents} then details each major block: the coordinator,
the two subagents, and the skills. Section~\ref{sec:exav} details the symbolic
core -- ontology, knowledge graph, and the \querier -- emphasising how it
departs from the rigid \exao workflow and should be read as a novel design.
Section~\ref{sec:arch} describes the comparison architectures.

\subsection{Architecture Overview}
\label{sec:arch}
\begin{figure*}[t]
  \centering
  \includegraphics[width=\textwidth]{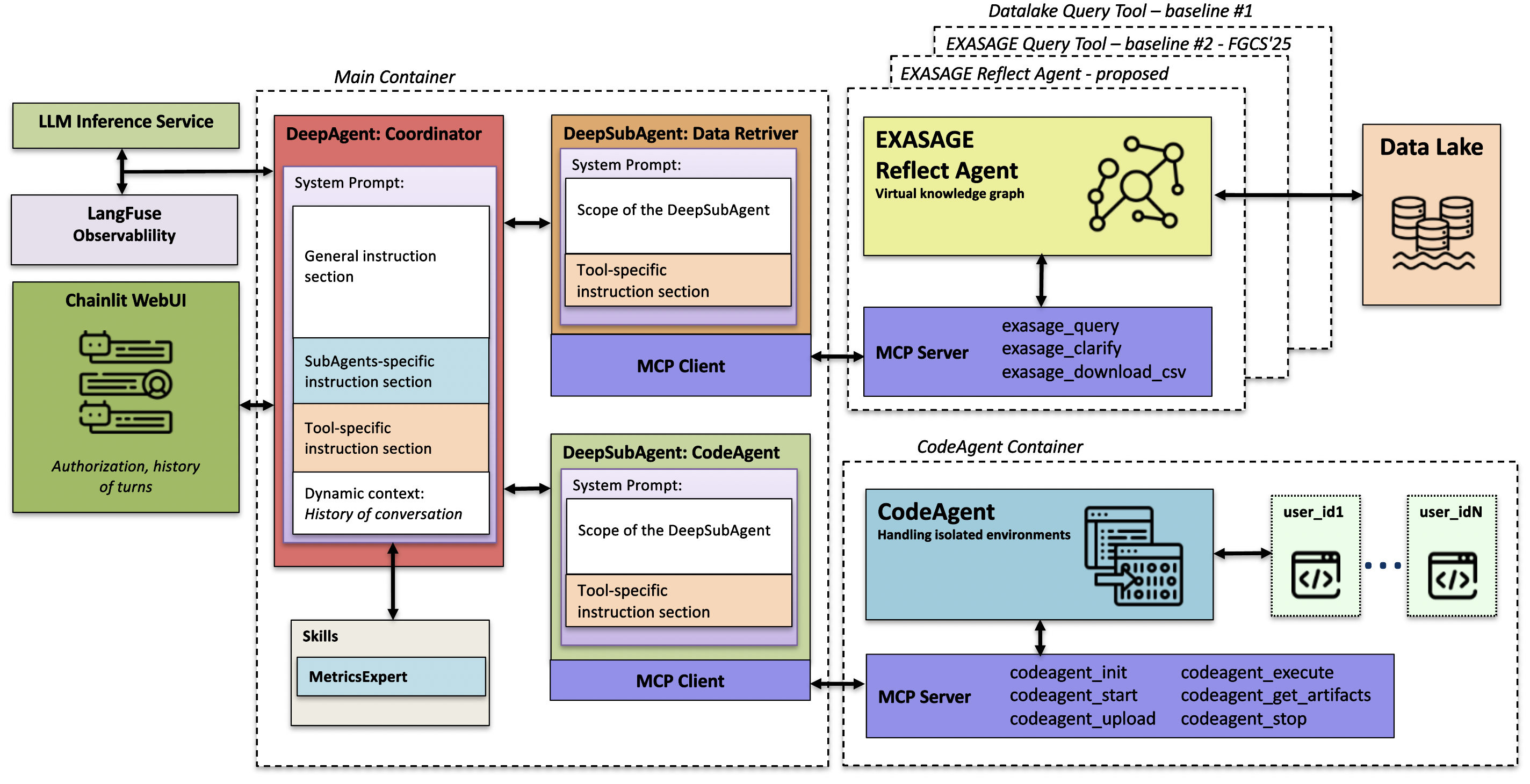}
    \caption{Reference architecture of the proposed \sys. 
    }
  \label{fig:arch}
\end{figure*}

Figure~\ref{fig:arch} shows the architecture of \sys. The system is organised as a
set of containerised services communicating exclusively through the Model Context
Protocol (MCP)~\cite{mcp_spec} over a FastMCP transport~\cite{fastmcp}, following
the MCP separation between a \emph{host} (the coordinator) and \emph{servers} (the
specialist tools). Three containers carry the logic. The \emph{main container}
holds the deep-agent \textbf{coordinator}, its skills, and the two
subagents, each of which owns an MCP client. The \emph{DataRetriever container} holds
the symbolic retrieval subagent -- the \textbf{\querier} -- which grounds all
telemetry access in an ontology-constrained Virtual Knowledge Graph built on demand
over the ODA Data Lake. The \emph{\codeagent container} holds the
\textbf{\codeagent}, which runs analysis and visualisation code inside
per-user, isolated sandboxes. Around these sit three supporting services: a local
LLM inference service (vLLM~\cite{kwon2023vllm}) shared by all reasoning
components, LangFuse~\cite{langfuse} for end-to-end observability, and a Chainlit
web interface that handles user authorisation and conversation history.

\paragraph{Flow of information}
A request enters through the web interface and reaches the coordinator, which is
responsible only for planning and delegation and deliberately holds a small context
window. The coordinator decomposes a multi-step request into an ordered task list;
for a task such as ``retrieve the average GPU temperature and plot it'' it first
dispatches the retrieval step -- as a \emph{natural-language} sub-request -- to the
\querier subagent through the MCP client/server pair, optionally consulting the
MetricsExpert skill to disambiguate metric names beforehand. The \querier resolves
the request into an ontology-validated SPARQL query over a per-query VKG
materialised from the Data Lake and returns compact, typed results, inline or as a
CSV reference for large outputs. The coordinator then forwards the retrieved data
to the \codeagent subagent, which composes and runs Python in an isolated
sandbox and returns the resulting artefacts (plots, statistics). Finally, the
coordinator presents the artifacts to the user with a concise summary. Every LLM
generation, tool call, and subagent dispatch along this path is recorded as a
nested trace in LangFuse.

\paragraph{Symbolic separation} \label{sec:symb}
The coordinator and subagents reason and communicate (Fig.~\ref{fig:arch}) in natural language and exchange typed, bounded results; on the data retrieval, the \querier never lets the LLM touch raw telemetry: all data access is mediated by the ontology-constrained VKG, and every generated SPARQL query is checked against the ontology \emph{before} execution (Section~\ref{sec:exav}) -- the \emph{domain-semantic} contract of Section~\ref{sec:soa}. The non-symbolic baseline (A3, Section~\ref{sec:arch}) is precisely the same architecture with this boundary removed: its retrieval subagent reads the raw datalake directly, so the LLM must discover schemas and infer joins itself.

\subsection{Components}\label{sec:subagents}
We now detail each major block of Fig.~\ref{fig:arch}. The agent layer is built on
LangGraph, which provides the low-level primitives -- nodes, edges, state graphs,
and checkpointers -- for modelling agent behaviour as a directed graph in which
each node invokes an LLM, executes a tool, or routes onward. On top of it, we use
the DeepAgents abstraction, whose factory instantiates a runnable agent from a
target LLM, a governing system prompt, a set of directly invoked tools, a 
list of subagents, a filesystem backend, and markdown skill files, abstracting away
the graph wiring, state management, and subagent lifecycle. Each subagent runs its
own MCP client and connects to a dedicated MCP server, so that a subagent is aware
only of its own tool suite; because communication is mediated by MCP rather than
direct references, subagents cannot reach into one another's containers, which
prevents accidental data leakage and lets individual retrieval servers scale
independently of the coordinator.

\paragraph{Coordinator (deep agent)}
The coordinator is the top-level deep agent that receives every request and
orchestrates the response. Its behaviour is governed by a layered system prompt
whose structure is shared across the compared configurations, with
configuration-specific sections appended as needed. A shared base layer establishes
its role as an orchestrator rather than an executor, states global rules such as
multi-step decomposition and a bounded dispatch limit, and outlines the
routing--dispatch--present workflow. A configuration-specific layer refines this
with routing rules for the active subagents. A tool-description layer exposes the
subagent descriptions used to decide where to dispatch a task. Finally, a runtime
layer injects, before each turn, the current user and conversation identifiers and
a sliding window over recent turns; this context is managed explicitly to prevent
the LangGraph checkpointer from replaying the full, unbounded message history and
overflowing the LLM context window. Given a multi-step request, the coordinator
decomposes it into an ordered task list, dispatches the retrieval step to the
\querier (optionally consulting the MetricsExpert skill first), forwards the
returned dataset to the \codeagent for analysis or plotting, and presents the
resulting artefacts with a concise summary. Every dispatch is tracked with a named
counter; if a subagent exhausts its internal retries, the coordinator records a
single failed dispatch and may re-attempt it, refining the request from the
subagent's error report, up to a bounded number of times before reporting the
outcome and terminating.

\paragraph{The \querier{} (symbolic retrieval sub-agent)}
The \querier is the symbolic heart of the \dr. Operating as a reflect loop over the
ontology-grounded VKG service (Section~\ref{sec:exav}), it retrieves telemetry
\emph{exclusively} through that service, behind an MCP server exposing three tools:
\texttt{exasage\_query} (retrieve data for a natural-language request),
\texttt{exasage\_clarify} (resume the pipeline from the appropriate stage when
disambiguation is needed), and \texttt{exasage\_download\_csv} (export large
results as a CSV file). This makes the pipeline \emph{interactive} and
\emph{stateful}, while guaranteeing that no raw table or fabricated column name
ever reaches the reasoning layer.

\paragraph{The \codeagent{} (analysis sub-agent)}
Absent from all prior EXASAGE work, the \codeagent transforms the telemetry returned by the \querier into visualisations and statistical analyses. To bound the risk of executing arbitrary code, it adds a further isolation layer based on nested containerisation: when a task requires execution, the \codeagent MCP server spawns a per-user sandbox container that has no outbound network connectivity, runs under the gVisor syscall-level runtime~\cite{gvisor} to intercept and filter system calls, exposes only two mounted directories for input and output, and is pre-provisioned with common data-science libraries (\texttt{pandas}, \texttt{numpy}, \texttt{matplotlib}, \texttt{scipy}, \texttt{scikit-learn}, \texttt{seaborn}, \texttt{statsmodels}, \texttt{openpyxl}). 

Sandboxes are provisioned per user, and a static auditor flags medium- and high-risk imports (e.g.\ \texttt{subprocess}, \texttt{socket}, \texttt{ctypes}) before execution. The component exposes an MCP tool set covering the sandbox lifecycle: \texttt{codeagent\_init} creates the container, \texttt{codeagent\_start} launches the executor, \texttt{codeagent\_upload} transfers a base64-encoded payload into the input directory without granting the subagent direct filesystem access, \texttt{codeagent\_execute} runs the code and returns the output, error logs, and the list of generated files, \texttt{codeagent\_get\_artifacts} retrieves outputs (such as a plot image), and \texttt{codeagent\_stop} shuts down the sandbox to release resources. 

A typical flow initialises the sandbox, uploads retrieved data, generates and executes a plotting script, and returns the artefacts to the coordinator for presentation. This component turns the system from a question-answerer into a genuine \emph{data analyst} and opens the door to integrating what-if analyses of the retrieved data for future work.

\paragraph{Skill: MetricsExpert}
Unlike a subagent, a skill has no LLM context, tool set, or reasoning loop; it is a markdown reference injected into the coordinator's system prompt as background knowledge, avoiding the overhead of spawning a subagent for what is essentially a lookup. Routing itself is handled by the coordinator's system prompt (the configuration-specific delegation layer above), so the only skill in \sys is \emph{MetricsExpert}.

It includes all M100 data-collection plugins (Ganglia, IPMI, Job Table, Nagios, Schneider, SLURM, Vertiv, Weather) with their metrics, units, sampling periods, and semantic descriptions, and is consulted by the coordinator when dispatching, and by the \querier during metric extraction, to resolve a natural-language description such as ``overall node power consumption'' or ``GPU temperature'' to an exact metric name (e.g.\ \texttt{total\_power} on the IPMI plugin, or \texttt{Gpu0\_gpu\_temp}). This approach raises query accuracy, which matters because the LLM generation step is prone to hallucination without proper resolution of metrics.

\paragraph{Observability}
Every request carries user, trace, and conversation identifiers threaded through
the coordinator, subagents, and individual tool calls, so that each interaction is
fully traceable in LangFuse~\cite{langfuse}. Each request produces a trace, that records every LLM generation with its
input/output tokens, latency, and model name; tool executions appear as distinct
spans with their arguments and response summaries; and subagent dispatches are
preserved as nested sub-traces, exposing the full coordinator $\rightarrow$
subagent $\rightarrow$ tool sequence. This yields the per-request execution graphs
and token accounting used for the stress-tests of Section~\ref{sec:results}.
Observability is not merely an engineering convenience: in a trustworthy analytics
setting it lets an operator audit \emph{how} an answer was produced, which
subagent was invoked, which SPARQL query was generated and validated, and where a
hallucinated construct was rejected -- rather than taking the final answer on
faith.

\subsection{The Symbolic Core: Ontology, Knowledge Graph, and the \querier}
\label{sec:exav}
The defining property of the approach is that the agent reasons \emph{about} the
data but acts \emph{only} through this symbolic layer, via two mechanisms detailed
below. First, \emph{relationships are pre-connected}: VKG construction materialises
the relevant telemetry into a single connected RDF graph in which job--node,
node--sensor--reading, and node--rack--room relations are already edges, so a
complex question is answered by \emph{one} ontology-validated traversal rather than
by the LLM inferring joins across queries. Second, \emph{access is
conformance-checked}: a deterministic validator rejects any query that names a
class, property, or domain/range combination absent from the ontology before it
executes. These two mechanisms are what make the \querier a novel design rather
than an increment over \exao.

\subsubsection{An ontology-constrained knowledge graph}
\label{sec:ontology}
Data access is governed by a domain ontology for ODA. Because this ontology is the
schema against which every generated query is validated, its expressiveness bounds
both \emph{what} the agent can answer and \emph{which} hallucinations the validator
can catch; it is itself a contribution of the use case. The first public ODA
ontology underpinning \exao~\cite{khan_fgcs_exasage, ontology_graphsys25} was adequate
for compute-node-centric queries but exhibited structural limitations that directly
constrained the analytics agent: it modelled only
\texttt{data center}$\rightarrow$\texttt{HPCSystem}$\rightarrow$\texttt{Rack}$\rightarrow$\texttt{ComputeNode}
(no rooms, no generic containment); kept all classes at one level, mixing physical
and logical entities; had no representation for cooling or power equipment (so the
Vertiv, Schneider, and Logics plugins were unreachable); modelled no external
sustainability metrics; and offered weak, mostly-direct relationships without
inferred containment for multi-hop reasoning.

\begin{figure}
    \centering
    \includegraphics[width=0.5\linewidth]{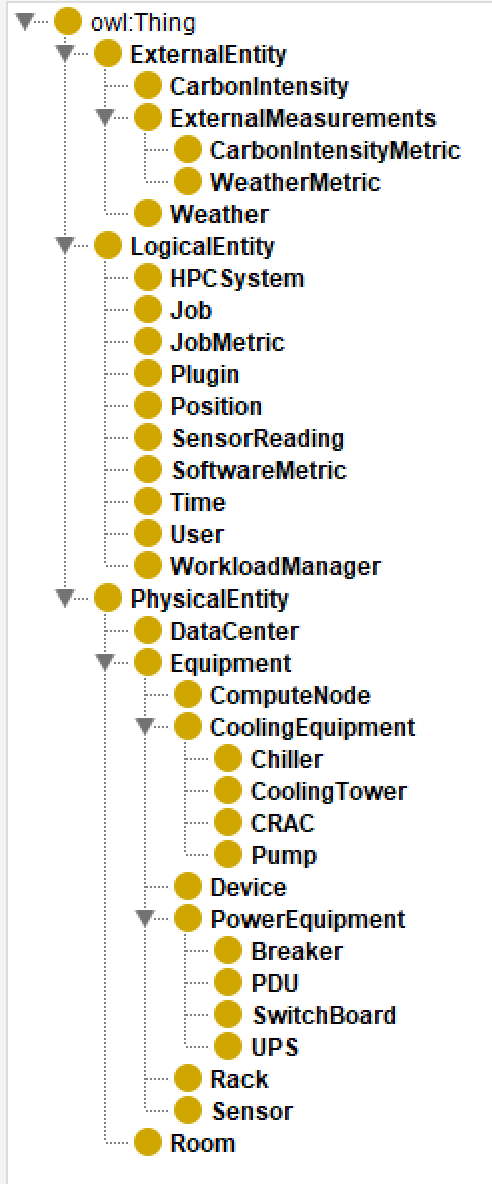}
    \caption{High-level class hierarchy of the revised ODA ontology, showing the three top-level classes: PhysicalEntity, LogicalEntity, and ExternalEntity, and their respective subclasses.}
    \label{fig:ontology_hierarchy}
\end{figure}

\begin{figure}[t]
  \centering
  \includegraphics[width=\linewidth]{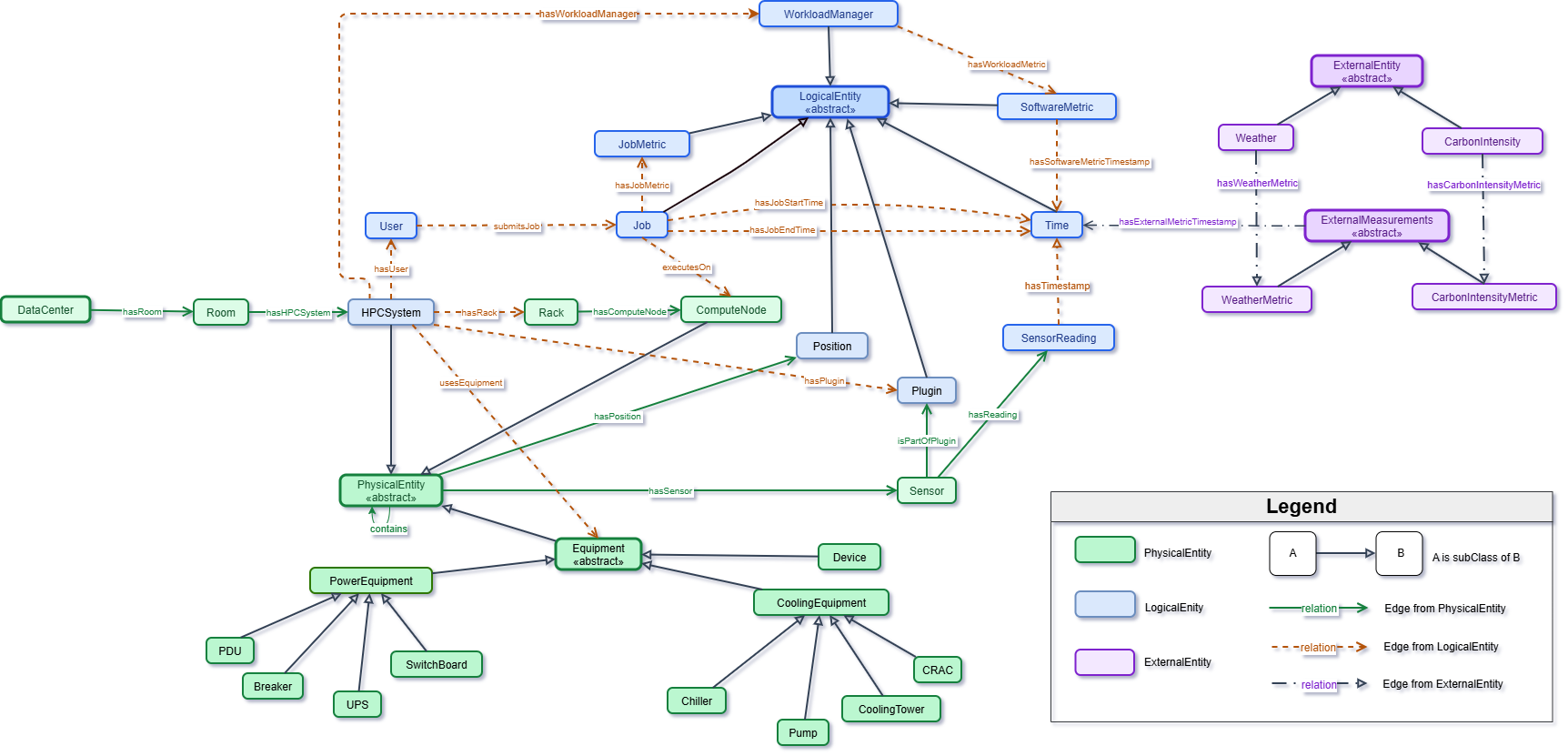}
  \caption{The revised ODA ontology that grounds the \querier. A layered
  Physical/Logical/External design, a full facility hierarchy with transitive
  containment, and expanded power, cooling, software, and sustainability coverage
  replace the compute-node-centric structure of the prior
  version~\cite{ontology_graphsys25}.}
  \label{fig:ontology}
\end{figure}

The revised ontology (Fig.~\ref{fig:ontology}) addresses each limitation and is
substantially larger and more expressive -- 308 axioms, 36 classes, 23 object
properties, and 55 data properties -- restructured into three disjoint top-level
classes: PhysicalEntity, LogicalEntity, and ExternalEntity. This separation
ensures a clear distinction between infrastructure, operational, and environmental
concepts. A high-level view of this organisation is shown in Fig.~\ref{fig:ontology_hierarchy}. 

\emph{PhysicalEntity} represents the tangible infrastructure of the data center.
It includes structural and hardware components such as data center, Room, Rack,
ComputeNode, and Equipment, including a breakdown of power system components (UPS, PDU, Breaker, SwitchBoard)
and cooling system components (Chiller, Pump, CoolingTower, CRAC). These entities are
connected through a transitive \texttt{contains} relation, which is defined
strictly within the PhysicalEntity hierarchy and enables multi-hop reasoning. 
However, the containment structure is not a single linear hierarchy but a set of
branching sub-hierarchies across different physical components. Additionally, 
Sensors can be attached to any PhysicalEntity, providing a consistent 
monitoring interface across all physical infrastructure components.

\emph{LogicalEntity} captures operational, execution and scheduling, and software-level
abstractions of the system. This includes the HPCSystem as the central cluster-level abstraction,
along with workload management components (WorkloadManager), job execution entities
(Job), and associated metrics (JobMetric and SoftwareMetric). It also encompasses
temporal modelling through Time, enabling consistent representation of system state
during execution. Overall, this layer models system behavior independently of physical
infrastructure, enabling reasoning over execution, performance, and operational state.

\emph{ExternalEntity} models environmental and external contextual factors that
influence system operation. It includes Weather conditions and CarbonIntensity
signals, enabling the integration of sustainability-aware analytics into the same
queryable ontology. This allows the system to reason about environmental impact
alongside internal system state.

For the agent, these changes are not cosmetic: the layered separation and
transitive containment let the deterministic ontology-conformance validator reason
about class hierarchies when rejecting hallucinated constructs; the expanded
equipment and software classes widen the set of answerable queries (and the plugins
the \querier can reach); and the sustainability entities make carbon-aware
analytics expressible. In short, the revised ontology is the schema that makes
\emph{symbolic separation} both broad and strict for this use case.

\subsubsection{From \exao to the \querier: a redesigned, validated pipeline}
\label{sec:querier-pipeline}
\begin{figure}[t]
  \centering
  \includegraphics[width=\columnwidth]{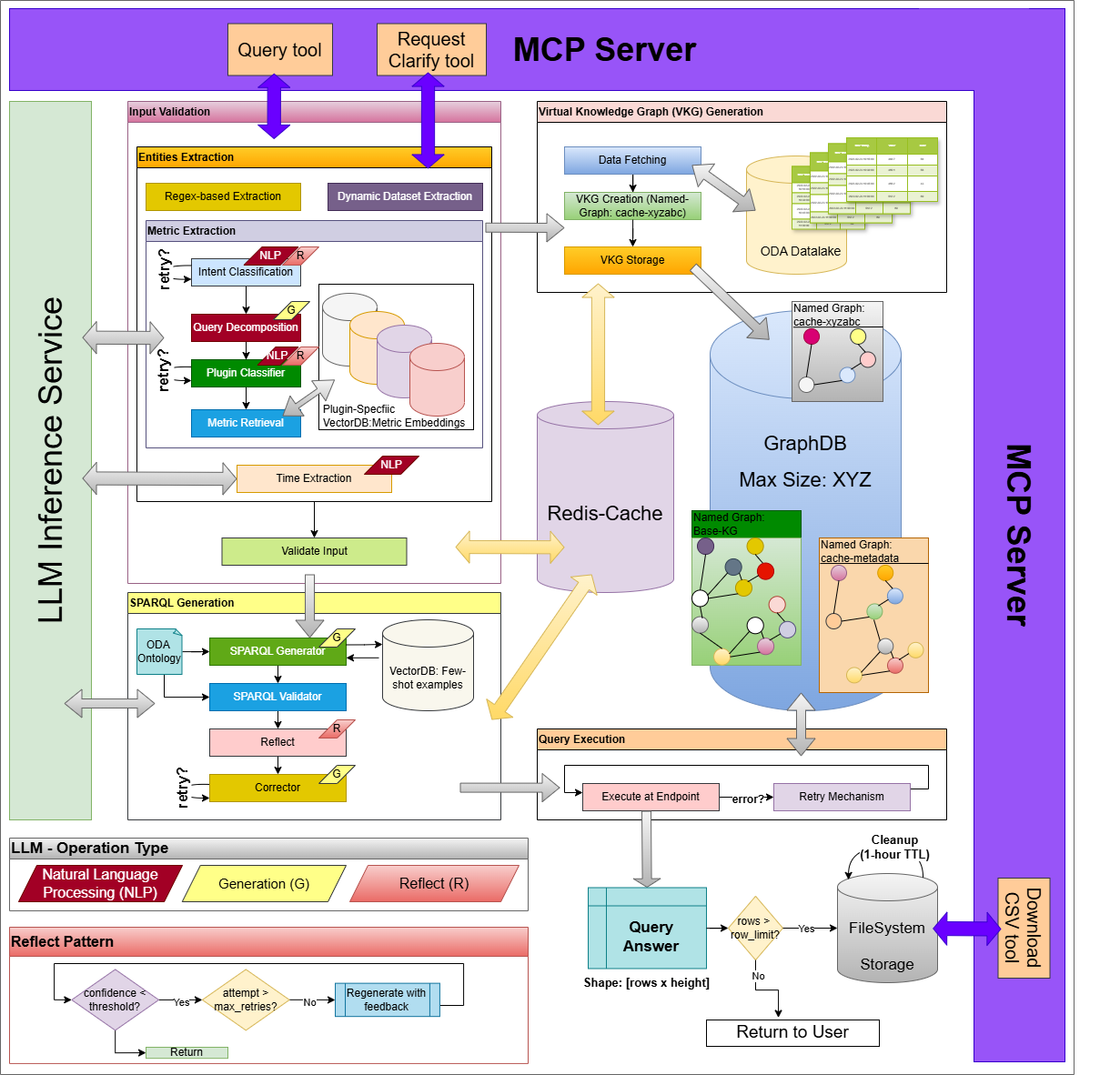}
  \caption{Block diagram of the \querier pipeline: input validation
  (entity, time, and multi-stage metric extraction), ontology- and
  retrieval-augmented SPARQL \& VKG generation with deterministic
  ontology-conformance validation and a bounded self-correction loop, and query
  execution, with a Redis-backed cache and named-graph VKG store.}
  \label{fig:exasage_v2_figure}
\end{figure}

Figure \ref{fig:exasage_v2_figure} summarizes the redesigned \querier{} pipeline, which replaces the brittle regex-based \exao with an LLM-driven, ontology-validated approach. The pipeline exposes three external tools through its MCP server: a query tool for handling natural-language requests and serving as the entry point, a clarify tool to resolve runtime ambiguities within user requests, and a download-CSV tool to export results exceeding the interface row limit (to avoid a crash) for external analysis. Behind these interfaces, the system processes requests across three distinct stages:

\begin{itemize}
    \item \emph{Input Validation:} A hybrid phase combining an (LLM-driven) extraction step for complex metric and temporal data with a (symbolic), regex-based extraction step for standard categories. These outputs are then verified by a strict, rule-based (symbolic) logical validation step.
    \item  \emph{SPARQL \& VKG Generation:} A bridging phase that uses an (LLM-driven) generator to translate natural language into (symbolic) SPARQL code, while the VKG generation step maps data directly to the (symbolic) schema of the underlying ontology.
    \item \emph{Query Execution:} A (purely deterministic) phase that runs the finalized, structured SPARQL queries directly on the materialised VKGs stored in the graph database.
\end{itemize}

The \querier{} leverages a shared LLM inference server across three distinct operating modes: Natural Language Processing (semantic analysis), Generation (structured JSON/SPARQL outputs), and Reflection (self-correction governed by an empirical confidence threshold, \(t_{\text{ref}} = 0.7\)).

The pipeline integrates two distinct storage layers to handle runtime state persistence and performance optimization independently: (1) a Redis cache manages pipeline checkpoints and session restores during external interruptions---most notably when the pipeline pauses to wait for a user's response to a clarification request. Each request tracks a unique identifier mapping to a state tuple, $S = (\texttt{stage}, \texttt{status}, \mathcal{I})$, where \texttt{stage} is the current pipeline stage, \texttt{status} is the execution state (\texttt{running}, \texttt{waiting-for-clarification}, \texttt{completed}, or \texttt{failed}), and $\mathcal{I}$ represents the intermediate artifacts (extracted entities, metrics, SPARQL, and VKG components); (2) a Named-Graph VKG store operates as a semantic cache that persists generated VKG artifacts so that the system can reuse them for future queries instead of rebuilding the same graph from scratch.

\paragraph{Stage 1: Input Validation}
This stage interprets the natural-language user request, extracts the referenced
telemetry entities, and verifies that the request is complete, requesting
clarification when a mandatory entity or parameter is missing. Fixed
infrastructure entities (node, rack, system, data center) are extracted reliably using rule-based regex extraction;
the two components that were most brittle in \exao -- temporal and metric
extraction -- are redesigned. Time extraction is implemented as a single-step LLM 
task leveraging few-shot in-context learning. The model identifies temporal expressions,
normalizes them, and emits standardized ISO-8601 start and end timestamps; this enables 
the pipeline to seamlessly handle colloquial phrasings such as ``the first day of September''
or ``during the previous week'' that \exao previously dropped. Metric extraction is harder 
because operational telemetry is heterogeneous and abstract terms must be grounded without requiring
the user to know schema-encoded metric names: the same word (``power'') maps to
different metrics depending on context (node-level power via IPMI, GPU power via
Ganglia, facility power via the electrical-panel plugins). 

We formulate metric extraction as a multi-stage semantic grounding problem combining LLM reasoning with embedding retrieval, expressed as $F(q) = (M \circ P \circ D \circ I)(q)$, where $I$, $D$, $P$, and $M$ denote intent classification, query decomposition, plugin classification, and metric retrieval respectively. The pipeline produces a set of grounded tuples $\mathcal{O} = \{(q_i, \pi_i, m_i)\}_{i=1}^{n}$, where each tuple maps a sub-query $q_i$ to its corresponding plugin $\pi_i$ and metric $m_i$, and $n$ denotes the number of decomposed sub-queries.

\emph{Intent classification $I(q)\rightarrow(i,c,p)$} determines whether the query is metric-related ($i$) and, if so, its cardinality ($c \in \{\texttt{single}, \texttt{multi}\}$), with confidence $p$. Non-metric queries are handled by an early exit path returning a non-metric response. Low-confidence predictions ($p < t_{\text{intent}}$) are refined using a reflection mechanism, iterated until $p \ge t_{\text{intent}}=t_{\text{ref}}$ or a fixed retry budget is exhausted. 

\emph{Query decomposition $D$} splits a multi-metric request into a set of context-preserving sub-queries $\mathcal{Q} = \{q_1,\dots,q_k\}$, each retaining the original entity, temporal, and scope context while isolating a single metric intent.

\emph{Plugin classification $P(q_i)\rightarrow(\pi_i,p_i)$} assigns a telemetry plugin $\pi_i$ to each sub-query with confidence $p_i$ under the same reflection mechanism, accepting only predictions with $p_i \ge t_{\text{plugin}}=t_{\text{ref}}$ and discarding sub-queries that fail to meet this threshold.

\emph{Metric retrieval $M(q_i, \pi_i)$} grounds each validated query--plugin pair against a per-plugin vector database of metric embeddings. Candidates are ranked by cosine similarity $s_{ij}$, with top score $s_i^\ast = \max_j s_{ij}$.

A decision function $\delta(s_i^\ast)$ determines the retrieval outcome:
\[
\delta(s_i^\ast) =
\begin{cases}
\texttt{fail}, & s_i^\ast < 0.5,\\[2pt]
\texttt{clarification}, & 0.5 \le s_i^\ast < 0.7,\\[2pt]
\texttt{accept}, & s_i^\ast \ge 0.7.
\end{cases}
\]

Only \texttt{accept} outcomes contribute to the output set $\mathcal{O}$. \texttt{clarification} outcomes suspend execution, return candidate metrics to the user, and resume upon receiving feedback, while \texttt{fail} outcomes terminate the current request and prompt the user to reformulate the query.

Algorithm~\ref{alg:metric} summarises the complete metric extraction pipeline.
Compared with the previous \exao implementation, this design resolves the failure
mode in which abstract metric references could not be grounded to database
metrics, while also supporting multi-metric requests. For example, the query
``gpu \emph{and} cpu utilisation for job $X$'' is decomposed into two sub-queries
and grounded to the metrics
\texttt{[Gpu3\_gpu\_utilization, cpu\_num]}.


\begin{algorithm}[t]
\caption{\querier Metric extraction}
\label{alg:metric}
\begin{algorithmic}[1]
\Require User query $q$
\Ensure Grounded query--plugin--metric tuples $\mathcal{O}$

\State $(i,c,p) \leftarrow I(q)$
\If{$i = \texttt{non\_metric}$}
    \State \Return non-metric response
\EndIf

\State retry $\leftarrow 0$

\While{$p < t_{\text{intent}}$ \textbf{and} $retry < max\_retries$}
    \State Apply reflection for self-correction
    \State $(i,c,p) \leftarrow I(q)$
    \State retry $\leftarrow retry + 1$
\EndWhile

\If{$p < t_{\text{intent}}$}
    \State \Return failure
\EndIf

\State $\mathcal{Q} \leftarrow D(q,c)$; \ \ $\mathcal{O} \leftarrow \emptyset$

\ForAll{$q_i \in \mathcal{Q}$}

    \State $(\pi_i,p_i) \leftarrow P(q_i)$; \ \ retry $\leftarrow 0$

    \While{$p_i < t_{\text{plugin}}$ \textbf{and} $retry < max\_retries$}
        \State Apply reflection to refine plugin selection
        \State $(\pi_i,p_i) \leftarrow P(q_i)$; \ \ retry $\leftarrow retry + 1$
    \EndWhile

    \If{$p_i < t_{\text{plugin}}$}
        \State \textbf{continue}
    \EndIf

    \State $\{(m_{ij},s_{ij})\} \leftarrow M(q_i,\pi_i)$; \ \ $s_i^\ast \leftarrow \max_j s_{ij}$

    \If{$s_i^\ast \ge 0.7$}
        \State $m_i \leftarrow \arg\max_j s_{ij}$; \ \ $\mathcal{O} \leftarrow \mathcal{O}\cup\{(q_i,\pi_i,m_i)\}$

    \ElsIf{$0.5 \le s_i^\ast < 0.7$}
        \State \Return clarification request

    \Else
        \State \Return failure request
    \EndIf

\EndFor

\State \Return $\mathcal{O}$

\end{algorithmic}
\end{algorithm}


\paragraph{Stage 2: SPARQL and VKG Generation}

The validated output of Stage~1 drives the generation of two artefacts: a query-specific VKG and the SPARQL query executed over it. Rather than materialising the full graph, the system constructs only the sub-graph required for the current request, following the schema of the revised ODA ontology (Section~\ref{sec:ontology}), and stores it as a named graph. VKG construction is deterministic and non-recoverable: any failure during graph assembly aborts the request.

\emph{VKG Generation} is a computationally intensive process, primarily due to the construction of large-scale RDF expansions over event-driven and time-series data. Job entities scale with user submissions $N_j$, while metric ingestion is driven by continuous sampling at frequency $f_s$, yielding $N_r \approx T \cdot f_s$ observations; since each observation expands into approximately four RDF triples (as defined by the revised ODA ontology in Section~\ref{sec:ontology}), the construction cost is dominated by metric processing in practice.

To bound re-computation, VKG generation is cached using a hash of extracted entities. Each cached entry is stored as a named graph and indexed in a dedicated \texttt{cache-metadata} graph containing creation time, last access time, size, and access-control attributes. Before constructing a new VKG, the system checks for an existing compatible entry; if found, it is reused and its access timestamp updated. Otherwise, a new VKG is generated and stored. Cached graphs are evicted using a least-recently-used policy when capacity is exceeded. VKG construction adopts the Polars-based batching and N-Triples serialisation optimisations introduced in our VKG-chatbot preprint~\cite{khan2025datacenteriottelemetry}, reducing latency while maintaining per-query graph storage in the order of a few MiB.

Our implementation adopts a custom VKG materialisation strategy tailored to the revised ODA ontology. Established VKG tools such as Ontop~\cite{calvanese2017ontop} were originally designed for relational databases but have since been extended to support heterogeneous data sources, including Parquet-based and non-relational systems commonly used in HPC environments. A comparative evaluation against such frameworks is deferred to future work.

\emph{SPARQL generation} is ontology-and context-augmented. The prompt incorporates extracted metrics, temporal constraints, and ontology traversal paths associated with each plugin, which act as structural constraints that reduce the space of admissible graph patterns and mitigate invalid joins. Furthermore, query generation is supported by retrieval-augmented few-shot prompting: previously curated question--SPARQL pairs are embedded in a dedicated vector database, and the top-$k$ most similar examples are retrieved into the prompt.

Generated queries are first processed by a \emph{deterministic ontology-conformance validator}. The validator extracts admissible classes, object properties, and data properties from the ontology along with their domain, range, and datatype constraints. It parses the generated SPARQL triple patterns, verifies that all referenced entities exist, and performs ontology-guided type inference using subclass, domain, and range relations. It also enforces structural consistency, including valid variable binding in projections, correct filter expressions, and connected graph patterns. Queries that violate these constraints are rejected before execution.

Queries that pass \emph{conformance validation} are then evaluated for semantic completeness using a reasoning LLM model. The model produces a structured report including a confidence score $p$ and diagnostic signals such as missing projections, projection quality issues, and grouping violations. If $p < 0.9$, a bounded repair loop is triggered. This loop uses diagnostic feedback to determine whether corrections are required, and applies a constrained instruction-following LLM restricted to modifying only SELECT and GROUP BY clauses. WHERE patterns and filter logic remain immutable. Repaired queries are optionally revalidated before execution to ensure ontology compliance.

Finally, SPARQL projections are rewritten to map internal identifiers to human-readable labels for downstream presentation (instead of internal KG URIs).

\paragraph{Stage 3: Query Execution}

The validated query is executed against the graph database SPARQL endpoint. Syntax-based execution failures trigger a bounded retry loop in which feedback is returned to the SPARQL generation module for regeneration. The system allows up to a maximum number of retries before terminating with failure.

\begin{figure}[t]
  \centering
  \includegraphics[width=\columnwidth]{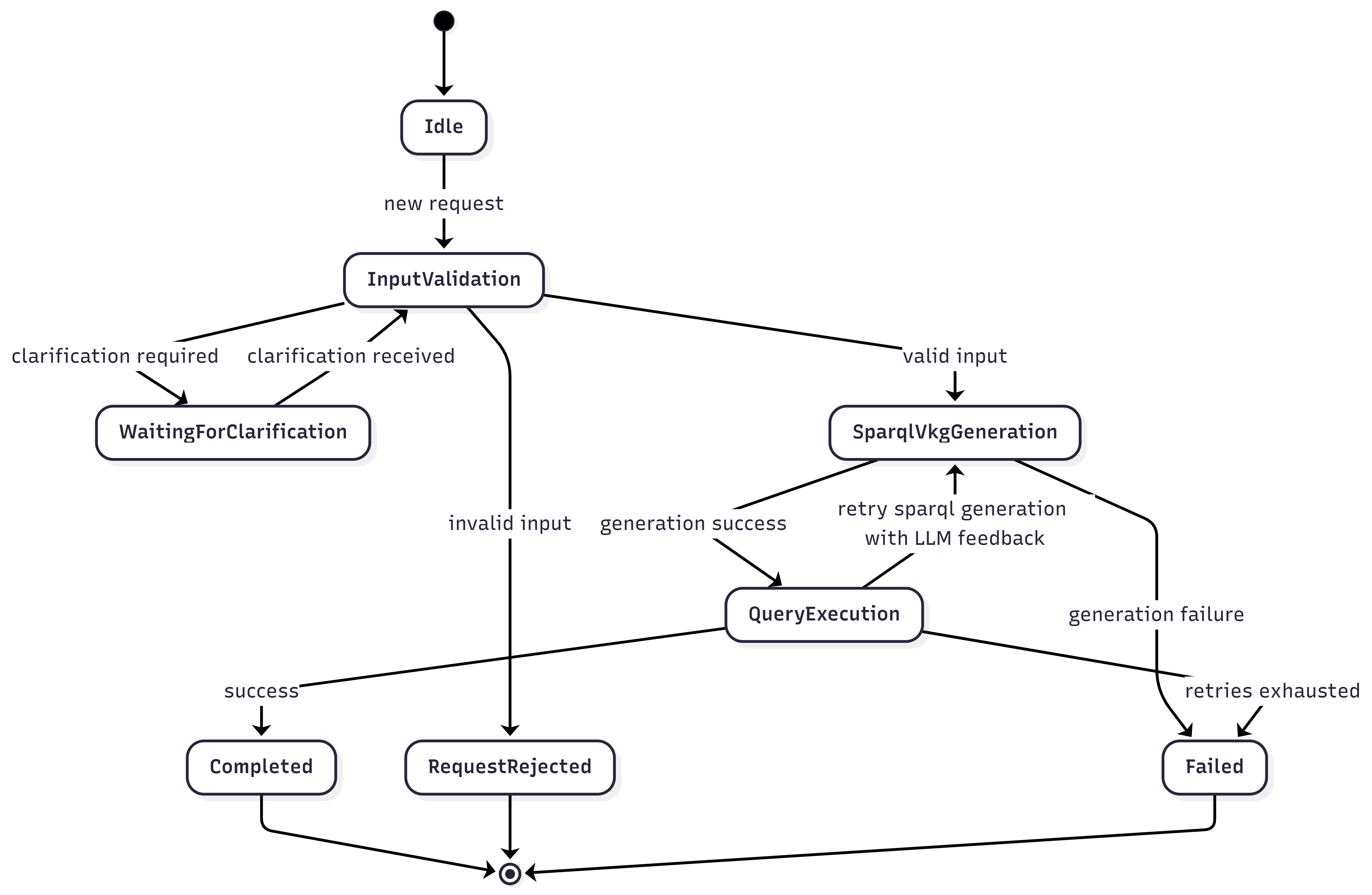}
  \caption{\querier state machine.
  }
  \label{fig:state_machine}
\end{figure}

\paragraph{System lifecycle}
Figure~\ref{fig:state_machine} summarises the request lifecycle as a state
machine. Execution begins in \emph{Idle}; a new request moves to \emph{Input
Validation}, from which three paths are possible: an incomplete or ambiguous
request enters \emph{Waiting for Clarification} and, once answered, returns to
input validation; an invalid request moves to the terminal \emph{Request Rejected}
state; and a valid request proceeds to \emph{SPARQL and VKG Generation}. Within
that stage, the two artefacts are treated differently: VKG construction is a
deterministic workflow with no self-correction, so any error in building the
query-specific graph transitions the request directly to the terminal
\emph{Failed} state; SPARQL generation, by contrast, runs the self-correction loop,
regenerating on failed ontology-conformance validation or execution until success
or exhaustion of the retry budget, at which point the request also reaches
\emph{Failed}. On successful generation, the request moves to \emph{Query
Execution} and then to the terminal \emph{Completed} state. All three terminal
states converge on the final state, ending the lifecycle.

The result is qualitatively different from \exao: interactive (it can pause and
ask the user), self-correcting (reflection plus deterministic ontology-conformance
validation), multi-sensor aware, and able to reach previously unsupported plugins.
Within the approach, the \querier is the symbolic anchor that makes the whole agent
trustworthy.

\section{Experimental Results}\label{sec:results}
In this section we evaluate the proposed \sys{} on a real data center-telemetry corpus. We compare three configurations: the proposed \sys{} (\dr{}
$\rightarrow$ \querier), the SoA \exao{}~\cite{khan_fgcs_exasage}, and the non-symbolic
\deepanalyst{} (\dr{} $\rightarrow$ \datalaketool) on two different sets of queries: (i) 14 end-to-end data analysis tasks and (ii) data retrieval only. The end-to-end data analysis queries are newly proposed, while the data retrieval ones are an extended version of the original ten queries archetypes extended to 75 unique queries. 
On the first end-to-end data analysis query dataset we tested two modern self-deployed open-weight LLMs: GPT-OSS-120B dense model and Qwen3.6-35B-A3B MoE model.

To isolate the effect of symbolic separation, we compare three configurations that
share the same coordinator, skills, LLM, and hardware and differ only in the
backend the \dr{} calls over MCP:
\begin{description}[leftmargin=0em, itemsep=2pt]
  \item[NSA -- \sys{} (proposed).] 
  The \dr{} calls the \querier{}
  (Section~\ref{sec:exav}): a reflect loop over the VKG that resolves ambiguous
  metric names, requests clarification on ambiguity, splits large time ranges to
  avoid timeouts, and returns results inline or as CSV. It exposes
  \texttt{exasage\_query}, \texttt{exasage\_clarify}, and
  \texttt{exasage\_download\_csv}.
  \item[SoTA\#1 -- \exao \cite{khan_fgcs_exasage} (SoA baseline).] 
  The \dr{} calls the
  published EXASAGE workflow \cite{khan_fgcs_exasage} over the same knowledge graph.
  Lacking the reflect loop, it cannot resolve metric names or clarify; it needs
  exact metric names (from the MetricsExpert skill) and strict
  \texttt{[YYYY-MM-DD HH:MM:SS]} timestamps, and a malformed input fails
  immediately. It exposes the same tool interface as the previous case.
  \item[SoTA\#2 -- \deepanalyst{} (non-symbolic ablation).] The \dr{} calls the
  \datalaketool, which bypasses the knowledge graph and queries the raw datalake
  directly through Datalake
  Query API and helper functions, with \emph{no} ontology validation. It can list
  plugins and metrics and run a targeted query, returning inline JSON, but has no
  reflect loop, clarification, or CSV export. The LLM must discover schemas itself
  and \emph{infer the relationships} to join multiple sources at query time, so SoTA\#2
  tests whether pre-connecting relations in the VKG (NSA) avoids that cross-call
  relational hallucination.
\end{description}

All experiments use M100~ExaData~\cite{m100nature}
(Parquet partitions); the Base-KG stores static M100 metadata (spatial layout,
rack configuration, node locations). All runs are traced in LangFuse~\cite{langfuse}, providing per-query execution graphs and token
accounting for reproducibility and audit.

Inference ran on an NVIDIA H100 80\,GB GPU under
vLLM~\cite{kwon2023vllm} with a maximum context length of 256k tokens; graph store GraphDB-Free; retrieval reads Parquet via Polars.

We evaluate the approach using two newly created sets of queries: (i) \emph{end-to-end}, a set of 14 queries (Appendix~A, Table~1) where retrieval is followed by analysis
and visualisation, so that both the symbolic contract and the deep-agent
orchestration are exercised together; (ii) a retrieval-level query dataset composed of 75 queries from recent HPC ODA
studies~\cite{ruad_martin, sencan_analyzing_gpu_utilization,
antici_online_job_failure, ANTICI_online_algorithm_for_job_power_prediction,
graafe} (Available at \href{https://gitlab.com/ecs-lab/multi-agent-system-public/exasage-reflect-agent-public}{GitLab} repository): for each study we translated its data pipeline into the questions an
analytics system must answer, an LLM drafted them, and five HPC researchers
rephrased them. This dataset itself includes a single instance of each of the ten query archetypes used in \cite{khan_fgcs_exasage} to compose the evaluated 1K queries.

\subsection{End-to-end data analytics task comparison}

We use a benchmark of
14 end-to-end analytics queries (Appendix~A Table~1) spanning the
spectrum of HPC operational tasks: each query requires retrieving the correct
telemetry \emph{and} producing a requested visualisation or statistic, so that a
trial is a full complete success only if both the data-retrieval and the code-execution
stages succeed. We run the three architectures
NSA, SoTA\#1, SoTA\#2, each with two open-weight LLMs served
locally -- Qwen3.6-35B-A3B~\cite{qwen3} and GPT-OSS-120B~\cite{gptoss} -- giving
$3\times2\times14 = 84$ end-to-end trials.

\begin{table}[htb]
\centering
\scalebox{0.85}{
\begin{tabular}{@{}llccc@{}}
\toprule
\textbf{Configuration} & \textbf{LLM} & \textbf{Data retrieval} & \textbf{Code exec.} & \textbf{Both} \\
\midrule
\querier & Qwen3.6-35B-A3B        & 13 & 12 & \textbf{12} \\
\querier & GPT-OSS-120B       &  5 &  4 & \textbf{3}  \\
\exao & Qwen3.6-35B-A3B           &  1 &  1 & \textbf{0}  \\
\exao & GPT-OSS-120B          &  1 &  0 & \textbf{0}  \\
\datalaketool & Qwen3.6-35B-A3B     &  5 &  2 & \textbf{2}  \\
\datalaketool & GPT-OSS-120B    &  8 &  7 & \textbf{6}  \\
\bottomrule
\end{tabular}
}
\caption{End-to-end results success rate
}
\label{tab:overall}
\end{table}

Table~\ref{tab:overall} reports for each tested configuration the number of queries that completed only the data retrieval part, only the code executor part, or both. It must be noted that in this table we count as completed also the cases in which the provided answers were wrong. 
\begin{table*}[h!t]
\centering
\caption{Per-query comparison, two best configurations. Status: C = correct, H = hallucinated, I = incomplete,
F = failed. \emph{Coord/Data/Code} partition the internal tokens across the
coordinator, data-retrieval, and code-execution components; \emph{Out} is the
share of generated (completion) tokens in the total tokens; \emph{Rt} is the number of failed
dispatch retries.}
\label{tab:per_question}
\scriptsize
\setlength{\tabcolsep}{4pt}
\begin{tabular}{l ccccccc | ccccccc}
\toprule
 & \multicolumn{7}{c}{\textbf{NSA -- \querier w. Qwen3.6-35B-A3B}} & \multicolumn{7}{c}{\textbf{SoTA\#2 -- \datalaketool w. GPT-OSS-120B}} \\
\cmidrule(lr){2-8} \cmidrule(lr){9-15}
\textbf{Q} & \textbf{St.} & \textbf{Tokens} & \textbf{Coord\%} & \textbf{Data\%} & \textbf{Code\%} & \textbf{Out\%} & \textbf{Rt}
           & \textbf{St.} & \textbf{Tokens} & \textbf{Coord\%} & \textbf{Data\%} & \textbf{Code\%} & \textbf{Out\%} & \textbf{Rt} \\
\midrule
Q1  & C & 44k   & 35.6 & 64.4 & 0.0  & 3.0 & 0 & H  & 92k   & 25.3 & 74.7 & 0.0  & 1.5  & 0  \\
Q2  & C & 149k  & 19.8 & 23.7 & 56.4 & 8.5 & 0 & C  & 561k  & 30.9 & 27.8 & 41.3 & 11.3 & 0  \\
Q3  & C & 360k  & 25.5 & 59.4 & 15.2 & 4.3 & 0 & C  & 462k  & 5.7  & 14.1 & 80.2 & 4.3  & 0  \\
Q4  & I & 589k  & 5.5  & 85.0 & 9.5  & 1.9 & 1 & I  & 2.50M & 4.2  & 6.6  & 89.2 & 0.7  & 0  \\
Q5  & C & 4.20M & 6.5  & 76.8 & 16.7 & 5.4 & 0 & F  & 630k  & 15.2 & 84.8 & 0.0  & 1.0  & 4  \\
Q6  & C & 4.50M & 15.1 & 54.8 & 30.2 & 5.6 & 1 & C  & 2.09M & 14.3 & 64.9 & 20.8 & 1.0  & 4  \\
Q7  & C & 240k  & 30.5 & 54.9 & 14.7 & 6.7 & 0 & F  & 114k  & 45.5 & 54.5 & 0.0  & 2.9  & 4  \\
Q8  & C & 3.64M & 0.9  & 97.6 & 1.5  & 1.0 & 0 & F  & 660k  & 1.1  & 98.9 & 0.0  & 1.0  & 4  \\
Q9  & C & 1.25M & 3.4  & 88.0 & 8.6  & 1.7 & 0 & F  & 167k  & 4.5  & 95.5 & 0.0  & 2.2  & 4  \\
Q10 & C & 100k  & 25.3 & 33.7 & 40.9 & 4.7 & 0 & C  & 713k  & 3.2  & 92.4 & 4.4  & 1.4  & 0  \\
Q11 & C & 45k   & 55.3 & 44.7 & 0.0  & 5.1 & 1 & FH & 669k  & 26.0 & 35.7 & 38.4 & 8.4  & 0  \\
Q12 & C & 80k   & 31.5 & 19.3 & 49.3 & 4.1 & 0 & FH & 1.18M & 1.2  & 98.8 & 0.0  & 0.9  & 0  \\
Q13 & F & 95k   & 8.6  & 91.4 & 0.0  & 2.9 & 6 & C  & 1.40M & 1.6  & 3.8  & 94.6 & 0.5  & 0  \\
Q14 & C & 510k  & 9.0  & 71.3 & 19.7 & 5.6 & 0 & F  & 1.46M & 96.0 & 4.0  & 0.0  & 1.1  & 14 \\
\bottomrule
\end{tabular}
\end{table*}
We can observe that both architecture and LLM model play a significant role. The proposed \sys{} and \querier{} when combined with Qwen3.6-35B-A3B achieves the highest success score of $86\%$. The two failed queries stop during the process as they reached the maximum context length --- the one that failed was the LLM inference server. In contrast SoTA\#1 configuration based on the \exao{} achieved only $7\%$ of success when using Qwen3.6-35B-A3B model ($0\%$ with GPT-OSS-120B one).

The single data retrieval and code exec. successes for \exao{} with Qwen3.6-35B-A3B occurred on two different queries. Manual inspection revealed that the one code-execution success was achieved illegitimately: the subagent, unable to retrieve data via EXASAGE, went beyond its tools by reading the filesystem and discovered a running MCP server of Datalake Query Tool and issued queries there. This cross-container escape further motivates the strict symbolic separation enforced by the proposed \sys{} architecture.

While the SoTA\#2 configuration, which does not rely on a neurosymbolic approach to query the data achieves a success score of $43\%$ with GPT-OSS-120B, but drops to $14\%$ with Qwen3.6-35B-A3B model. An analysis of provided answers shows that while for the proposed approach and SoTA\#1 all the successful queries are also correct, out of the 6 successful queries of the SoTA\#2 only 5 are correct.

This underline the importance of the symbolic separation design concept deep analyst agents as a mechanism to guarantee trustworthiness. Furthermore, it must be noted the significant gap between the $7\%$ of successfully queries and the $93.6\%$ of accuracy reported by the authors of EXASAGE in \cite{khan_fgcs_exasage}. We further investigate this gap: we conducted a second analysis on the data retrieval queries dataset in Section \ref{75queries}. The gap between the score achieved by \querier and the \exao answer to the RQ1 -- the newly presented \querier design is essential for accurate and trustworthy \sys agents.   

We now restrict to the most performing configuration, namely \querier with Qwen3.6-35B-A3B and SoTA\#2 with GPT-OSS-120B and we analyse their behaviour for the individual tested queries. This comparison answer RQ2. 

Table~\ref{tab:per_question} reports for each configuration tested: (i)~%
\emph{data-retrieval success} and \emph{code-execution success} (and their
conjunction, ``both successful''); (ii)~\emph{answer correctness}, with responses
labelled correct, hallucinated, incomplete, or failed; and (iii)~\emph{token
usage} which counts both input and output tokens, decomposed across the three internal components (coordinator, data
retrieval, code execution) as a cost/effort proxy. We also report the percentage of generated output tokens in the total token count. (iv) the number of failed try any agent or sub-agent faced during a query answer. All runs are traced in
LangFuse~\cite{langfuse}, providing per-query execution graphs and token
accounting for reproducibility and audit.

From it, we can notice that the complexity of each query resolution varied from tens of thousands of tokens to millions of tokens. Among these the generated output tokens are a small fraction of the total, ranging from 0.5\% to 11.3\%, confirming that multi-agent systems are predominantly context-processing workloads.

Overall, \querier{} with Qwen3.6-35B-A3B achieves a median of 300k tokens per successful query versus 713k tokens for \datalaketool{} with GPT-OSS-120B: a 2.4× token efficiency improvement, demonstrating that symbolic separation delivers both accuracy and cost savings.

In terms of retrials, the retry counts reveal a qualitative difference in error handling: \querier{} rarely retries beyond a single attempt, while \datalaketool{} repeatedly retries failing queries, burning millions of tokens, for example, 14 retries on Q14 alone without recovering. This suggests that the symbolic separation in \querier{} provides early detection of unsalvageable trajectories, avoiding wasteful computation.

Notably, \datalaketool{} produces hallucinated outputs on Q11 and Q12 (marked FH): answers that appear plausible but are factually wrong. This failure mode is more problematic than a hard failure, because a human observer might trust the output without manual verification. \querier{} produces zero hallucinated answers, further reinforcing the trustworthiness argument. Across the models generation steps, GPT-OSS-120B hallucinates the most: frequently inventing metric names and, at the agent level, terminating early, hallucinating missing data and mismanaging large results (on Q14 it looped 14 times for 1.46M tokens, 96\% in the coordinator with no proper output, whereas Qwen3.6-35B-A3B finished within 510k tokens). These results demonstrate that adherence to instructions and persistent reasoning matter more than raw parameter count across the models, while hallucination-prone model is prevented from returning correct data. 

Therefore, aggregating across all 14 data analytics tasks: \querier{} achieves 12 correct, 1 incomplete, 1 failed with a median of 0 retries. \datalaketool{} achieves 4 correct, 1 incomplete, 2 hallucinated, 5 failed with a median of 4 retries on failed queries. The correctness gap (86\% vs. 29\%) and the retry gap together demonstrate that the proposed symbolic separation improves both accuracy, trustworthiness, and cost predictability, demonstrating the \sys{} on Qwen3.6-35B-A3B is the best configuration. 

Since \exao{} reaches only $\sim$7\% here against the $93.6\%$ reported in~\cite{khan_fgcs_exasage}, we isolate the cause with a retrieval-level study. The full query is in Appendix~A.

\subsection{Data retrieval comparison}
\label{75queries}

Table~\ref{tab:retrieval} reports correct retrievals per backend for the formulated 75 queries on HPC ODA studies. The \exao{} answers 42/75 ($56\%$), which surpasses the \datalaketool accuracy, which answers only to 36/75 ($48\%$) questions. This confirms that \exao rigid, single-shot formulation is better than \datalaketool (which does not implement symbolic separation), and it is strong on templated queries, but degrades on free-form phrasing: a component breakdown (Table~\ref{tab:accuracy}) locates the loss in SPARQL generation (46.3\%) and final-answer composition (42.7\%) rather than entity extraction (63.4\%).  
\begin{table}[htb]
\centering

\footnotesize
\begin{tabular}{@{}lc@{}}
\toprule
\textbf{Data-retrieval backend} & \textbf{Correct retrievals} \\
\midrule
\datalaketool{}        & 36/75 \\
\exao{}      & 42/75 \\
\querier{} (proposed)  & \textbf{66/75} \\
\bottomrule
\end{tabular}
\caption{Retrieval-level study accuracy}
\label{tab:retrieval}
\end{table}

The \querier{} answers 66/75 ($88\%$), roughly doubling \exao{}; its residual failures trace to sub-agent query framing, SPARQL syntax, and out-of-context inputs rather than to grounding. These results confirm that decomposing NL → SPARQL into distinct stages with specialised sub-agents is what closes the gap and improves accuracy, not the agentic infrastructure itself.

\begin{table}[htb]
\centering
\footnotesize
\begin{tabular}{@{}lc@{}}
\toprule
\textbf{Component} & \textbf{Accuracy [\%]} \\
\midrule
Entity extraction                     & 63.4 \\
SPARQL query generation               & 46.3 \\
Virtual Knowledge Graph generation    & 59.8 \\
Final answer                          & 42.7 \\
\bottomrule
\end{tabular}
\caption{\exao{} Component-wise accuracy 
}
\label{tab:accuracy}
\end{table}

\subsection{The Deep Code Agent (analysis sub-agent)}. In this subsection, we evaluate the result provided by the \codeagent  which  demonstrates a powerful capability to autonomously extract, analyze, and visualize complex operational data from high-performance computing environments. We present (Appendix~B, Figure~1) the performance on three distinct analytical tasks (Q4, Q9, Q6) using the EXASAGE Reflect Agent with Qwen3.6-35B-A3B as the reasoning engine for data processing and visualization.

The successful generation of these plots each requiring distinct analytical techniques (time-series correlation, comparative profiling, and distribution analysis with outlier detection) demonstrates that analysis subagent is not merely a plotting tool but a cognitive extension of the EXASAGE. It enables the system to transform raw data into actionable insights, complete with contextual annotations and statistical summaries. This capability is particularly valuable in large-scale HPC environments where manual analysis is infeasible.

\section{Conclusion}\label{sec:concl}
We presented a neurosymbolic deep-agent \emph{approach} for trustworthy data
analytics over large-scale \emph{numerical} operational telemetry, built on the
principle of \emph{symbolic separation}: a generative agent may reason freely but
may act on data only through an ontology-constrained knowledge graph that
validates every access against the schema before execution. 
We validated the proposed \sys{} against two ablations on M100~ExaData across two open-weight LLMs. Symbolic grounding raised end-to-end task success from 6/14 for the strongest non-symbolic configuration to 12/14 for the \sys{} on Qwen3.6-35B-A3B, prevented silent data-integrity errors that no syntactic check catches, and did so
at roughly $2.4\times$ lower token cost on successful queries; a retrieval-level study
attributed the large accuracy gap to the published EXASAGE~\cite{khan_fgcs_exasage} result to the single-shot formulation of the \exao{} rather than to grounding, since the
\querier{} roughly doubled its correct retrievals. 
The results show that the proposed symbolic separation approach plays a key role in the creation of trustworthy data analyst agents. 
Future works will extend the \sys with a visual validator of generated artifacts,  extend the ontology and VKG approach to support stateful and incident queries, and extend the approach to other IoT and Industry 4.0 use-cases as well as validating it in real systems. 

\section*{Acknowledgment}
This research was supported by EuroHPC JU SEANERGYS (g.a.~101177590).

\bibliographystyle{IEEEtran}
\bibliography{references}

\end{document}


\appendices
\section{Benchmark Queries and answers}\label{app:extra}
This appendix lists the 14 end-to-end queries. 

\begin{table}[htb]
\centering
\caption{The 14 end-to-end analytics queries (Q1--Q14) used in the evaluation.
Each requires both correct retrieval and a correct visualisation/statistic.}
\label{tab:full_queries}
\footnotesize
\begin{tabular}{@{}p{0.03\columnwidth}p{0.88\columnwidth}@{}}
\toprule
\textbf{Id} & \textbf{Query} \\
\midrule
Q1  & How many nodes are there in M100? \\
Q2  & Retrieve the total power consumption for node 900 from 1 June 2022 20:00 to 23:59 and plot the time series. \\
Q3  & Retrieve GPU-0 utilisation for nodes 900 and 920 on 1 June 2022 from 08:00 to 15:00 and create a line plot comparing the nodes. \\
Q4  & Retrieve the inlet temperature for all nodes on 1 June 2022 between 08:00--09:00 and create a histogram of the temperature distribution. \\
Q5  & Retrieve ambient temperature for nodes 301 and 318 on 1 June 2022 from 08:00--15:00 and create a single comparative line plot. \\
Q6  & Retrieve temperature and power for node 900 on 1 June 2022 (full day) and create a dual-axis plot over time. \\
Q7  & Compute the coefficient of variation of GPU utilisation for jobs 3423336, 4835969, 3776044 over their whole duration and create a comparative bar chart. \\
Q8  & Get the total CPU and GPU hours consumed on the morning of 1 June 2022 and create a pie chart of the breakdown. \\
Q9  & For the top-3 longest single-node jobs on 1 June 2022, retrieve their power consumption, create a comparative line plot, and report min/max/avg power. \\
Q10 & Retrieve the average system-level CPU and GPU utilisation on 1 June 2022 between 15:00--16:00 and create a side-by-side bar chart. \\
Q11 & Retrieve the daily average temperature from the weather plugin for the first week of June 2022 and compare it with the same period of the previous month using a dual-axis plot. \\
Q12 & Get the number of jobs executed each day over 1--10 June 2022 and plot it over the 10 days. \\
Q13 & For node 900 on 1 June 2022, get ambient temperature, system CPU, GPU-0--4 utilisation, and total power from 12:00 to 14:00 and visualise a heatmap correlation matrix. \\
Q14 & Fetch power consumption and users on 1 June 2022 and create a box plot of the top-10 users by number of jobs and average power consumption. \\
\bottomrule
\end{tabular}
\end{table}

\section{Deep Code Agent Performance}
\label{app:plots}

\begin{figure}[ht]
    \centering

    \begin{subfigure}{\linewidth}
        \centering
        \includegraphics[width=\linewidth]{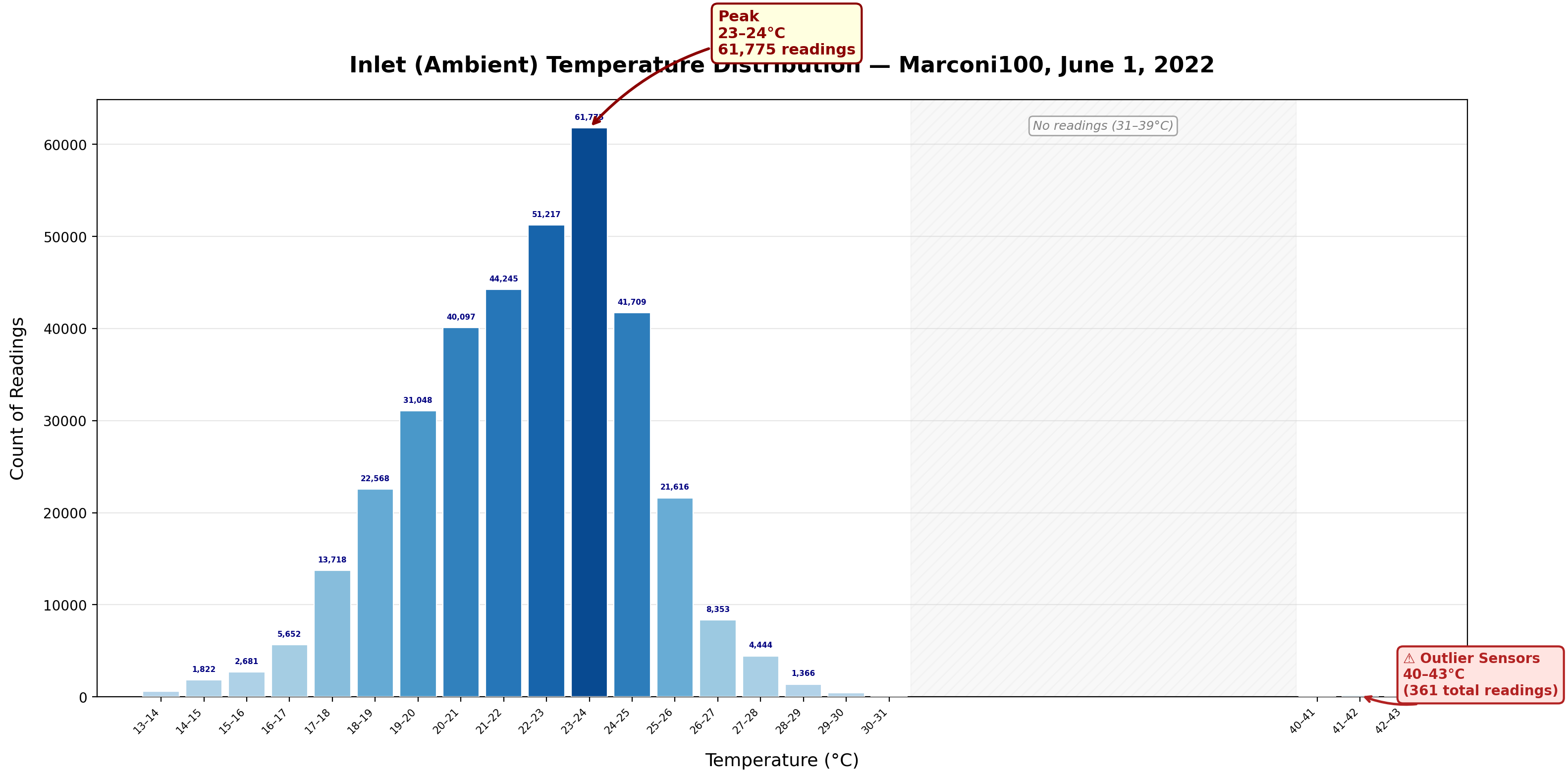}
        \caption{}
        \label{fig:hist}
    \end{subfigure}

    \vspace{0.5em}

    \begin{subfigure}{\linewidth}
        \centering
        \includegraphics[width=\linewidth]{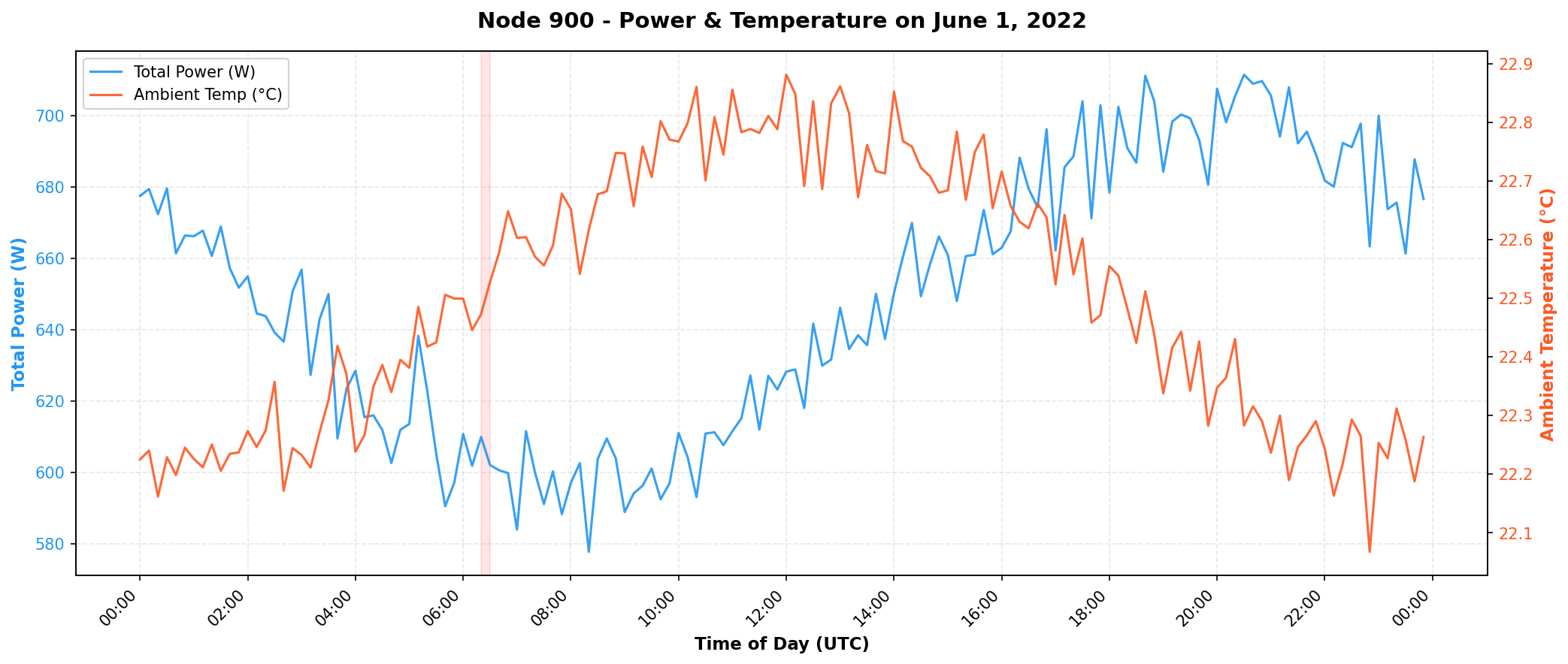}
        \caption{}
        \label{fig:temp}
    \end{subfigure}

    \vspace{0.5em}

    \begin{subfigure}{\linewidth}
        \centering
        \includegraphics[width=\linewidth]{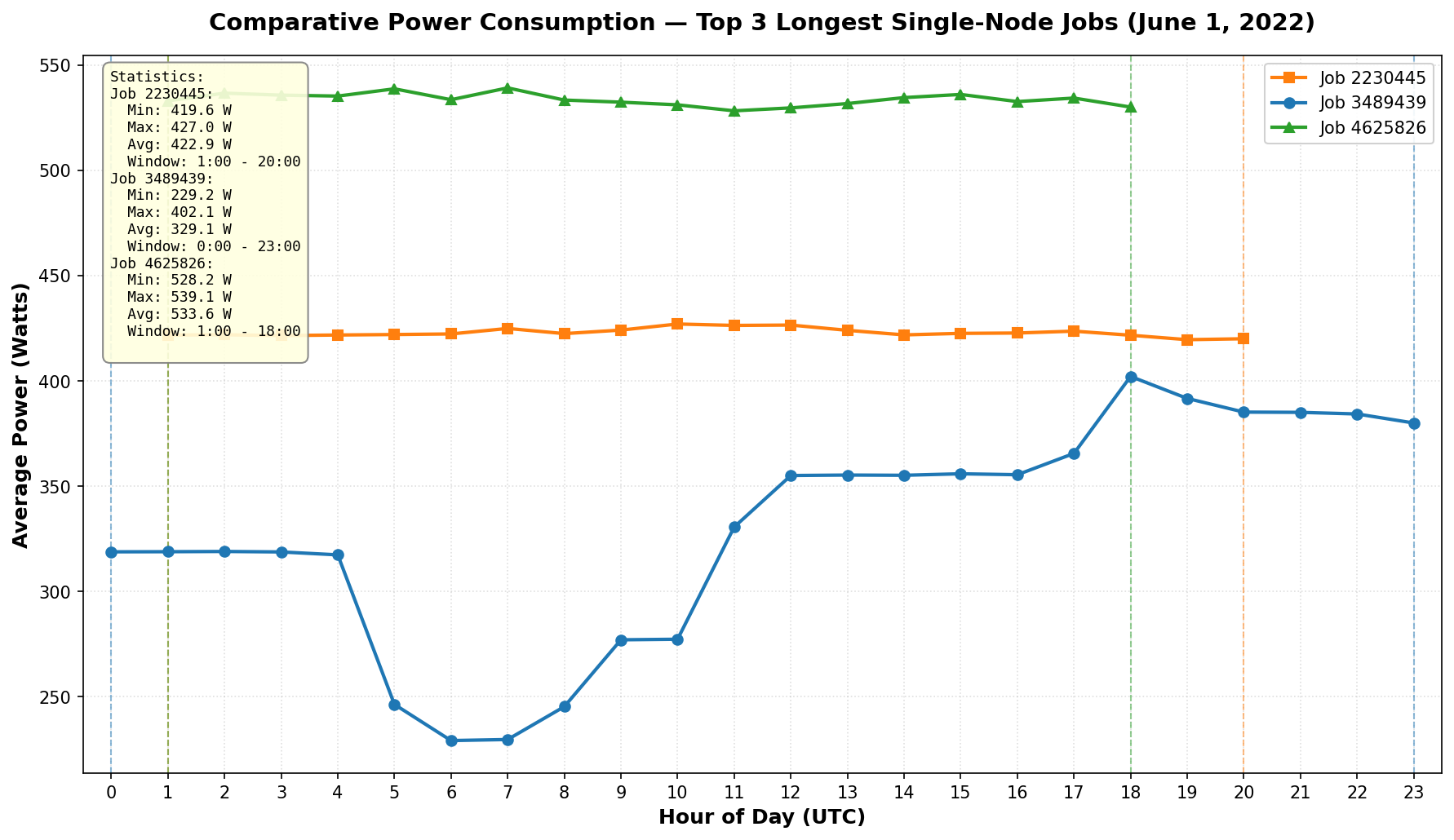}
        \caption{}
        \label{fig:power}
    \end{subfigure}

    \caption{Comprehensive Analysis of Marconi100 on June 1, 2022, generated by EXASAGE Reflect Agent with Qwen3.6-35B-A3B with Deep Code Agent.
    (a) Q4 - Inlet temperature distribution, highlighting peak range, absence of mid-range readings, and outlier sensors.
    (b) Q6 - Node 900 power and ambient temperature over time, revealing diurnal correlation and a transient event.
    (c) Q9 - Comparative power profiles of top three longest-running jobs, with annotated statistics.}
    \label{fig:all_plots}
\end{figure}

Figure~\ref{fig:all_plots} showcases the end-to-end analytical output generated entirely by the system. Subfigure~\ref{fig:hist} (Q4) presents the inlet temperature distribution across the Marconi100 system, revealing a pronounced peak at 23–24°C (61,775 readings), a complete absence of readings in the 31–39°C range, and a small cluster of outlier sensors registering 40–43°C (361 total readings). This histogram not only summarizes the thermal state but also enables anomaly detection. Subfigure~\ref{fig:temp} (Q6) displays the time-series correlation between total power consumption and ambient temperature for Node 900, highlighting a strong diurnal pattern and a transient event at around 06:30 UTC. Subfigure~\ref{fig:power} (Q9) compares the average power profiles of the top three longest-running single-node jobs, annotated with summary statistics (min, max, avg, and active window) for each, allowing operators to identify resource-intensive workloads and optimize scheduling.

Additional plots are available in the public repository at \href{https://gitlab.com/ecs-lab/multi-agent-system-public/exasage-reflect-agent-public}{GitLab}.

